\documentclass{article}
\usepackage{graphicx} 
\usepackage{appendix}
\usepackage{enumitem}
\usepackage{url}
\usepackage{tcolorbox}

\title{Playing With AI: How Do State-Of-The-Art Large Language Models Perform in the 1977 Text-Based Adventure Game Zork?}
\author{Berry Gerrits\\
\textit{University of Twente, Enschede, The Netherlands}\\
\texttt{b.gerrits@utwente.nl}}

\begin{document}

\maketitle

\begin{abstract}
In this positioning paper, we evaluate the problem-solving and reasoning capabilities of contemporary Large Language Models (LLMs) through their performance in \textit{Zork}, the seminal text-based adventure game first released in 1977. The game's dialogue-based structure provides a controlled environment for assessing how LLM-based chatbots interpret natural language descriptions and generate appropriate action sequences to succeed in the game. We test the performance of leading proprietary models---ChatGPT, Claude, and Gemini---under both minimal and detailed instructions, measuring game progress through achieved scores as the primary metric. Our results reveal that all tested models achieve less than 10\% completion on average, with even the best-performing model (Claude Opus 4.5) reaching only approximately 75 out of 350 possible points. Notably, providing detailed game instructions offers no improvement, nor does enabling ``extended thinking". Qualitative analysis of the models' reasoning processes reveals fundamental limitations: repeated unsuccessful actions suggesting an inability to reflect on one's own thinking, inconsistent persistence of strategies, and failure to learn from previous attempts despite access to conversation history. These findings suggest substantial limitations in current LLMs' metacognitive abilities and problem-solving capabilities within the domain of text-based games, raising questions about the nature and extent of their reasoning capabilities.
\end{abstract}

\section{Introduction}
Recent advances in Large Language Models (LLMs) have demonstrated impressive performance across a wide range of tasks, from code generation to idea generation, and from assistance in engineering problems to sparking creativity. These impressive capabilities have renewed optimistic predictions about the imminent arrival of Artificial General Intelligence (AGI) or human-level AI \cite{bubeck2023sparks}. However, such predictions echo earlier waves of AI optimism that ultimately failed to materialize. As Mitchell \cite{mitchell2021why} observes, the field of AI has cycled several times between periods of optimistic predictions and substantial investments (``AI spring'') and periods of disappointment, loss of confidence, and reduced funding (``AI winter''). From Herbert Simon's 1960 declaration that ``machines will be capable, within twenty years, of doing any work that a man can do'' \cite{simon1960new} to Marvin Minsky's forecast that problems of creating artificial intelligence would be ``substantially solved'' within a generation \cite{minsky1967computation}, the history of AI is marked by predictions that have often overestimated the intelligence of AI and underestimated the complexity of human intelligence.

One reason for these cycles is our limited understanding of the nature and complexity of intelligence itself \cite{mitchell2021why}. A critical but often overlooked aspect of AI is \textit{metacognition} — the ability to monitor and reflect on one's own thinking processes \cite{flavell1979metacognition, nelson1990metamemory}. As Mitchell \cite{mitchell2019artificial} argues, when humans struggle with a problem, they eventually recognize that repeating ineffective strategies is futile and change their course of action accordingly. This self-awareness represents cognitive abilities that current AI systems seem to lack. The absence of such metacognitive capabilities relates to what Rota \cite{rota1985} termed the ``barrier of meaning'' in AI: the fundamental challenge of how computational systems can genuinely understand concepts rather than merely utilizing a ``bag of tricks'' that only \textit{seems} intelligent. The bigger the bag, the more ``intelligent'' AI seems to be. While LLMs can generate text with remarkable fluency, they lack genuine understanding of the concepts they process \cite{bender2020climbing}. Without having interaction with the real-world like humans do, these systems struggle to adaptively reason when familiar patterns fail or training data falls short. For example, it does not take a human ---even an infant--- an excessive amount of training data (e.g., pictures of a cat) to be able to recognize a cat, to generalize the concept of a cat, and subsequently be able to reliable distinguish a cat from other objects (e.g., a rabbit). 

This raises questions about the extent to which LLMs can demonstrate genuine understanding and problem-solving capabilities. Text-based games provide a testbed for investigating this question as there is little to no room for deploying `tricks' to genuinely understand the game's mechanics as there is only sparse feedback. Rather, human players should leverage imagination, world knowledge, and adaptive reasoning to understand these fictional environments \cite{montfort2005twisty}.

\textit{Zork}, originally developed at MIT and released in 1977 (later commercially released by Infocom \cite{infocom1977zork} in 1980), represents a canonical example of interactive fiction that challenges players through pure textual interaction. The game presents players with descriptions of locations, objects, and situations entirely through natural language (e.g., ``You are standing in an open field west of a white house, with a boarded front door''), and players respond by typing commands in natural language (e.g., ``pick-up sword'' or ``go north''). Success in \textit{Zork} requires players to construct and continuously update mental models of spatial environments—imagining room layouts, remembering connections between locations, and tracking inventory—without any visual aids. Players must maintain a coherent state across long sequences of actions, recalling which puzzles they have solved, which items they have collected, and how different objects might interact. Moreover, the game provides sparse feedback through text responses, requiring players to learn from both successes and failures to refine their understanding of the game world's rules and logic.

What makes \textit{Zork} particularly relevant for testing LLMs is how it depends on distinctly human cognitive capacities. Human players leverage imagination to construct mental representations from textual descriptions, apply commonsense world knowledge to reason about object properties and causal relationships (understanding, for instance, that a lantern provides light, or that a sword can kill someone), and should continuously reflect on their own strategies, i.e., show some level of metacognition. As Montfort \cite{montfort2005twisty} states in his analysis of interactive fiction, these games reveal the sophisticated interplay between language comprehension, spatial reasoning, planning, and problem-solving that characterizes human intelligence.

This in stark contrast with the concept of Next Token Prediction (NTP) that has driven recent LLM successes. While an LLM might generate fluent text responses based on statistical patterns in its training data, \textit{Zork} arguably demands genuine comprehension: understanding that ``the door is locked'' implies the need for a key, that ``the room is dark'' requires a light source, or that certain puzzles have prerequisites that must be solved first. The question we address in this paper is whether contemporary LLMs can exhibit the flexible, goal-oriented problem-solving that text adventure games like \textit{Zork} demands, or whether their performance reveals a lack of genuine understanding that Mitchell \cite{mitchell2021why} identifies as characteristic of systems that have learned shortcuts rather than true conceptual understanding.

\section{Related Work}
Games have long served as benchmarks for AI capabilities, with particularly well-known examples including chess \cite{campbell2002deep} and Go \cite{silver2016mastering}. Most prominent successes involve games with well-defined state spaces and reward signals suitable for Reinforcement Learning (RL). Although impressive, Mitchell \cite{mitchell2021why} notes, what appears to be progress toward AGI may instead reflect Dreyfus's ``first-step fallacy''—the mistaken assumption that narrow intelligence exists on a continuum with general intelligence: “It was like claiming that the first monkey that climbed a tree was making progress towards landing on the moon'' \cite{dreyfus2012history}. Deep Blue's chess-playing achievement was indeed hailed as ``the first step of an AI revolution'' \cite{aron2016ai}, yet such narrow achievements have not translated into broader intelligence. 

Text-based adventure games present fundamentally different challenges than these well-structured games: they require natural language understanding, common sense reasoning about physical objects and spatial relationships, and the ability to form and test hypotheses about game mechanics \cite{cote2018textworld}. Recent work has explored LLM performance on text-based games with mixed results. Côté et al. \cite{cote2018textworld} introduced TextWorld as a platform for training RL agents on procedurally generated text games, though most approaches rely on fine-tuning rather than zero-shot reasoning. Ammanabrolu and Riedl \cite{ammanabrolu2019playing} demonstrated that incorporating commonsense knowledge graphs improves agent performance, suggesting that pure language models lack sufficient world knowledge. More recently, researchers have shown that while state-of-the-art LLMs can play some text-based games effectively, they struggle with games requiring long-term planning and spatial reasoning \cite{yao2023react}—precisely the capabilities central to \textit{Zork}.

\subsection{Reasoning in LLMs}
With respect to the general reasoning capabilities of LLMs, recent research has investigated various prompting and reasoning frameworks to enhance LLM problem-solving. Wei et al. \cite{wei2022chain} introduced Chain-of-Thought (CoT) prompting, showing that encouraging step-by-step reasoning improves performance on arithmetic and commonsense reasoning tasks. Yao et al. \cite{yao2023tree} extended this with Tree of Thoughts, enabling exploration of multiple reasoning paths. However, Kambhampati et al. \cite{kambhampati2024llms} argue that LLMs cannot generate executable plans and lack to ability to show self-critique, even when using CoT prompting. 

The abovementioned limitations relate to ``Moravec's paradox''—the counter-intuitive finding that tasks humans find difficult (chess, complex mathematics) are often easier for AI, while tasks humans find trivial (recognizing objects, navigating spaces, understanding context) remain challenging for machines \cite{moravec1988mind}. This paradox is particularly relevant to games like \textit{Zork}, where success depends less on computational power than on the kind of commonsense understanding that humans acquire naturally through interaction with the physical world.

The question of whether LLMs possess genuine understanding remains open for debate. Bender and Koller \cite{bender2020climbing} argue that language models trained only on linguistic patterns cannot acquire meaning without grounding in the real world. Marcus and Davis \cite{marcus2019rebooting} similarly contend that deep learning systems lack the robust, compositional understanding characteristic of human cognition. While some researchers argue that sufficiently large models may develop emergent understanding \cite{bubeck2023sparks}, critics maintain that pattern matching—however sophisticated—differs fundamentally from comprehension \cite{mitchell2021why}. 

Also Mitchell warns against the danger of ``wishful mnemonics'' \cite{mcdermott1976artificial}—using terms like ``understands,'' ``learns,'' or ``thinks'' to describe AI systems in ways that unconsciously shape our perception of their capabilities and lead us to overestimate their intelligence. 

\subsection{Metacognition in LLMs}

While metacognition has been extensively studied in human psychology \cite{flavell1979metacognition, schraw1994assessing}, relatively little work addresses metacognitive capabilities in AI systems. Some researchers have explored confidence calibration in neural networks \cite{guo2017calibration} and uncertainty estimation \cite{lakshminarayanan2017simple}, but these approaches differ fundamentally from human metacognition's adaptive, reflective qualities. 

Mitchell \cite{mitchell2019artificial, mitchell2021why} argues that current AI systems lack the self-monitoring necessary to recognize when they are ``stuck'' and need to change strategies—a limitation that may be particularly evident in open-ended problem-solving contexts like \textit{Zork}. Recent work by Yin et al. \cite{yin2023interpreting} on LLM self-evaluation found that models can assess their own outputs but also highlight that there is a considerable gap between the capabilities of the models and human profiency.

This connects to the broader challenge of common sense in AI, which Oren Etzioni has called ``the dark matter of artificial intelligence'' \cite{knight2018us}—a quality whose effects are visible everywhere but whose nature remains elusive. Common sense encompasses not just vast knowledge about the world, but the ability to use that knowledge flexibly to recognize situations, make predictions, and guide actions \cite{davis2015commonsense, levesque2017common}.

In this paper, we contribute to the debate by evaluating how state-of-the-art LLMs perform on \textit{Zork}, a game that requires players to build coherent mental representations, learn from feedback, and adaptively solve a series of puzzles. Instead of deploying an RL approach that trains agents through millions of episodes on a specific game, we examine LLMs in a zero-shot setting—mirroring how humans can pick up and play a new game without extensive prior training. By examining where and how these models succeed or fail, we aim to shed light on current limitations in LLMs' reasoning, spatial understanding, memory utilization, and metacognitive capabilities.

\section{Methodology}
We conduct our experiments using \textit{Zork I: The Great Underground Empire}, the first instalment of the commercially released Zork trilogy, originally developed at MIT in 1977 and later released by Infocom in 1980. While all games in the trilogy are considered challenging, the sequels (\textit{Zork II} and \textit{Zork III}) are arguably more difficult, making Zork I a suitable starting point for this research.
The game is accessed through a browser-based implementation \cite{iplayif}. A custom-built Python script connects the LLM and the game environment. The script captures the game's output after each move, forwards it to the LLM, receives the model's command, and inputs this command back into the game. The LLM has access to the full conversation of the current run.

We evaluate six LLM-based chatbots from three major providers:
\begin{itemize}[noitemsep]
    \item \textbf{Anthropic:} Claude Opus 4.5 and Claude Sonnet 4.5, with an additional variant of Opus 4.5 with `extended thinking' enabled.
    \item \textbf{OpenAI:} ChatGPT 5.2 in two configurations: `standard thinking' and `extended thinking'.
    \item \textbf{Google:} Gemini 3 with `thinking' enabled.
\end{itemize}

Each model plays the game independently, with no information \textit{explicitly} shared between runs or models. Each model does have access to previous chats which the model may or may not use in subsequent chats. Importantly, general-purpose LLMs have not been specifically trained or fine-tuned on \textit{Zork} or similar text adventure games (like RL approaches typically do to play games). While it is possible that descriptions of \textit{Zork} or walkthroughs exist within the models' training data, we do not provide any explicit game solutions in our prompts. This setup allows us to evaluate the models' ability for genuine problem-solving rather than memorization.

We distinguish between two instructions given by the initial prompt to the LLM. The \textit{basic prompt} provides minimal instruction, informing the model only that it is playing the game \textit{Zork}, that the user will provide game outputs, and that the model should respond exclusively with valid game commands. No background information about the game mechanics, objectives, or strategies are provided. The \textit{advanced prompt} offers comprehensive guidance derived from the original \textit{Zork I} instruction manual. This includes detailed information about movement and navigation commands, syntax, combat mechanics, communication with non-player characters, and general gameplay strategies. The prompt also provides context about the game's setting and objective: exploring the Great Underground Empire, collecting treasures, and depositing them in the trophy case. In both conditions, models are explicitly instructed not to use the \texttt{restart} command and are given the option to respond with ``I give up'' if they believe they are irreversibly stuck. Both prompts are provided in Appendix~\ref{prompts}.

Each model is tested under both prompts with five independent runs per model-prompt combination, resulting in a total of 40 experimental runs. Runs are conducted sequentially with fresh conversation contexts to prevent any carry-over effects between sessions (apart from access to chat history). For each run, we set the maximum number of moves to 500 and we collect the following data:
\begin{itemize}[noitemsep]
    \item \textbf{Number of moves.} The total count of commands issued before the run terminates. A run ends when the model issues a ``give up'' command, the player dies in the game, the game is successfully completed, or the maximum number of moves is reached.
    \item \textbf{Score.} \textit{Zork I} awards points as a rough indicator of progress in the game. That is, points are deducted when the player shows regression instead of progress. The maximum achievable score is 350 points. We record the score as an indicator of overall progress.
    \item \textbf{Conversation log.} The full transcript of game outputs and model responses is saved for further analysis.
\end{itemize}

\section{Results}\label{sec:results}
In this section, we quantitively discuss the performance of LLMs playing \textit{Zork}. In Section \ref{sec:discussion}, we further discuss the ``thinking'' and ``reasoning'' capabilities. 

\begin{figure}
    \centering
    \includegraphics[width=1.0\linewidth]{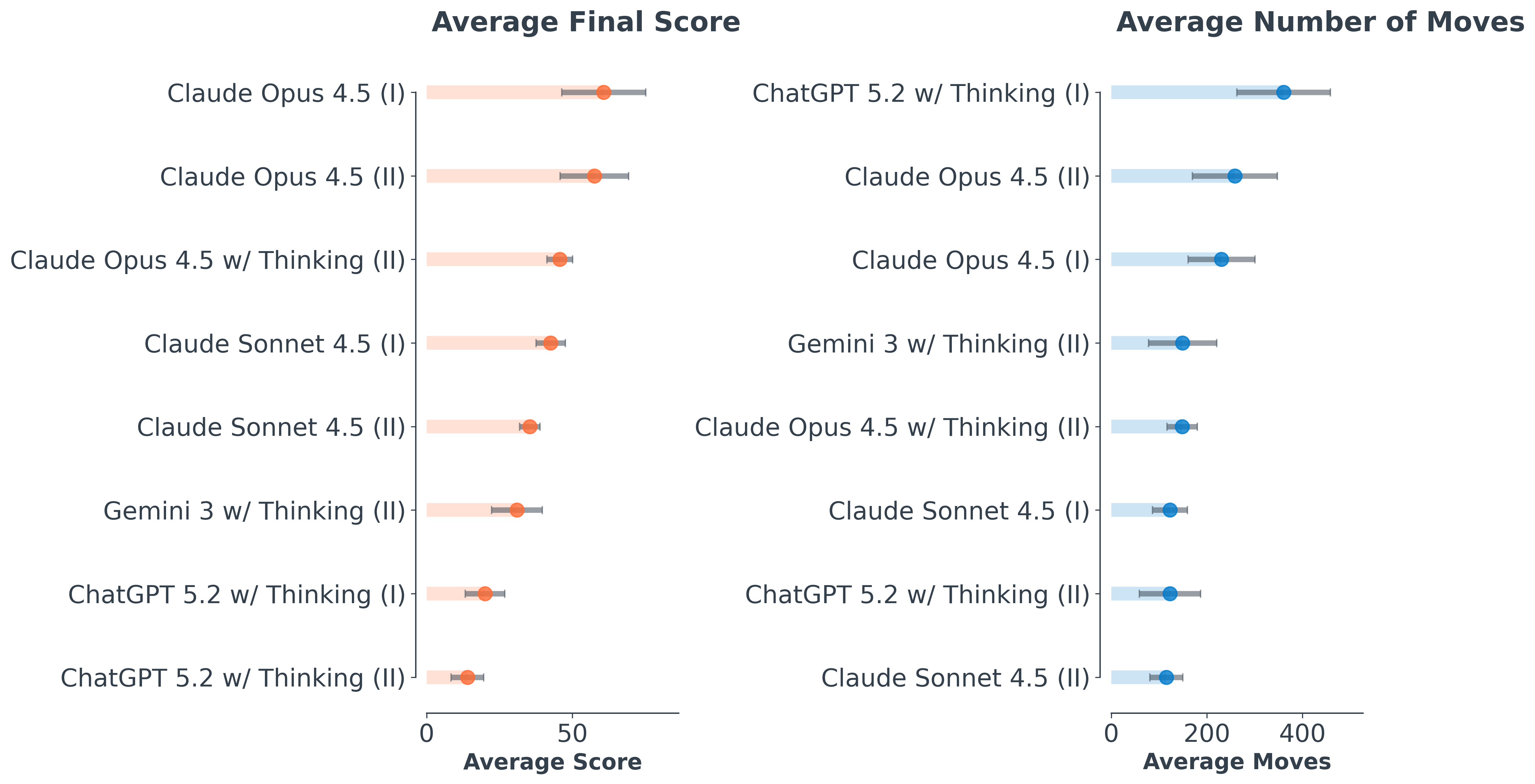}
    \caption{Left: Average number of points with standard error bars. Right: Average number of moves per game with standard errors bars. (I) denotes the basic prompt and (II) denotes the advanced prompt.}
    \label{fig:zork_analyses}
\end{figure}

Figure \ref{fig:zork_analyses} summarizes the key numerical results. On the left-hand side, the average final score is shown and reveals large differences between the studied LLM-based chatbots. First, note that all LLMs did not exceed a score of approximately 75 points, far below the required 350 points to finish the game. In other words, a maximum of $\sim$20\% progress was achieved, while on average less than $\sim$10\% of the game was completed. Second, the results suggest that Claude Opus 4.5 outperforms both Gemini 3 and ChatGPT 5.2, indicating that the models provided by Anthropic are more capable of playing the game. The difference between the variants of Claude appear minimal. Interestingly, there seems to be no difference between using the basic prompt (indicated by I) and the advanced prompt (indicated by II), while you may suspect that detailed instructions on the internal mechanics of the game, including multiple key gameplay insights, should improve performance. For human players, we suspect that access to a detailed manual has a significant impact on the ability to play the game \textit{Zork}. Lastly, the results suggest that there is no difference when `thinking' is enabled across all three providers. Although Figure \ref{fig:zork_analyses} does not show the difference between `non-thinking' and `thinking' for Gemini and ChatGPT, we found similar results as using Claude, i.e. minimal differences. Regarding Claude, Opus 4.5 seems to slightly outperform Sonnet 4.5, but `thinking' did not result in increased performance. These results suggest that what is labelled as 'thinking' does not translate to measurable improvements in gameplay performance, raising questions about whether these mechanisms capture the adaptive, metacognitive processes that enable human players to learn and improve at games like \textit{Zork}.

Although the number of moves does not say much about the ability to play the game, the right-hand side of Figure \ref{fig:zork_analyses} provides some insight in when the game ended. Recall that the game ends when (i) the maximum score is achieved, (ii) the player dies, or (iii) the LLM provides the command `I give up'. Interestingly, although ChatGPT performed the worst in terms of overall game progress, it refuses to give up, as indicated by the average number of moves until (ii) or (iii) are reached. Indeed, further inspection reveals that ChatGPT often finds itself in long series of similar commands resulting in no measurable progress (i.e., the player was stuck), hinting at no useful form of metacognition. To a lesser extent this also holds for the other LLMs, where ---due to further advancing in the game--- they found themselves more likely to be killed, either due to bad luck or due to improper game play (e.g., not killing a monster while its unarmed and stunned on the ground). Further analysis shows that only Claude issued the command `I give up' when it appeared to find itself in a vicious circle it could not escape. 

\section{Discussion} \label{sec:discussion}
In this section, a more qualitative discussion is presented on the reasoning and thinking capabilities of the LLMs under test. Starting at the worst performing model ---ChatGPT--- analysis of the chat history and reasoning explanations reveals distinct departures from human-like reasoning patterns. For example, in the beginning of the game there is a mailbox with a leaflet in it. All chatbots open the mailbox and read the leaflet to ``possibly obtain relevant information to play the game'', which seems reasonable. Although the leaflet does not provide much information to play the game, on several occasions ChatGPT returns to the mailbox and attempts to open it, only to find it empty. This does not suggest an intelligent form of thinking; any human player would remember that the mailbox was already opened. ChatGPT showed this kind of repetitive behaviour in multiple situations during the game. This also explains the large number of moves (due to repeated actions) whilst limited progress was made in the game, as discussed in Section \ref{sec:results}. 

Moreover, it seems that none of the chatbots learn from their previous runs even though they have access to the chat and conversation history. That is, mistakes made during one iteration are often repeated later in the same iteration or in following iterations, suggesting a lack of metacognition --- the ability to monitor and reflect on one's own thinking processes \cite{flavell1979metacognition, nelson1990metamemory}. Although no human players were included in the experimental setup, one would assume that a human player of any skill level would not repeatedly reconsider actions that have proved unsuccessful. The inability to learn from previous runs may relate to known LLM limitations in utilizing long contexts. Liu et al. \cite{liu2023lost} demonstrated that LLMs struggle to use information in the middle of long contexts, a phenomenon they term `lost in the middle.' In our \textit{Zork} experiments, relevant information about failed strategies from earlier in the conversation may be effectively invisible to the model when making decisions dozens of moves later.

Let us turn to the `best' performing LLM in our study: Claude Opus 4.5 and take a closer look at the thinking-enabled version. As an illustration, the thought process by Claude shown in Example 1 seems reasonable at first glance. 

\newtcolorbox{gameconvo}{
  colback=gray!5,
  colframe=gray!40,
  fonttitle=\bfseries,
  title=Example 1,
  left=3mm,
  right=3mm
}
\begin{gameconvo}
\textbf{Game output:} This is part of a maze of twisty little passages, all alike.\\
\textit{What command should I enter?}\\[0.5em]
\textcolor{blue}{Thought process: I'm in the maze and trying to find my way through. The maze in Zork I has a specific solution. Let me try to navigate it. I should keep dropping items to mark my path or try different directions systematically}\\[0.5em]
\texttt{> GO SOUTH}
\end{gameconvo}

The thought process suggests that Claude knows the solution to the maze in \textit{Zork I}, which could be the case if the solution was part of the training data. However, in the same thought it also states it should ``systematically try differ directions'', an approach that would be rendered moot if the solution were already known. If it actually does not know the solution to the maze, stating that the maze has a specific solution seems superfluous as \textit{any} maze has a specific solution. Moreover, it also suggest to ``keep dropping items to mark my path'', an approach that seems redundant as the chatbot can easily retrace its own steps based on the chat history. Furthermore, any human player would probably not deploy this strategy as these items are typically required to progress in the game. Indeed, one instance shows that Claude drops a lantern to mark its path, only to find itself later in the game entering a dark area without the ability to light the area with the lantern it had previously discarded. 

Example 2 shows the ability of Claude to reason about which item to drop based on the value it represents.

\newtcolorbox{gameconvo2}{
  colback=gray!5,
  colframe=gray!40,
  fonttitle=\bfseries,
  title=Example 2,
  left=3mm,
  right=3mm
}
\begin{gameconvo2}
\textbf{Game output:} Your load is too heavy.\\
\textit{What command should I enter?}\\[0.5em]
\textcolor{blue}{Thought process: I need to drop something to make room. The paper or leaflet might be droppable items since they're less valuable. Let me drop the leaflet first.}\\[0.5em]
\texttt{> DROP LEAFLET}
\end{gameconvo2}

This thought process seems reasonable enough to assume that the chatbot ``knows what it's doing'', but given the previous example and the relatively straightforward chain of thought, it might also suggest a form of pattern matching and merely utilizing a ``bag of tricks'' that only appears intelligent, recalling Rota's ``barrier of meaning'' in AI \cite{rota1985}. 

Example 3 (Page 11) shows what seems to be reasonable thinking but fails to deliver on that promise. Note that in this example, the player had previously already used the trap door, thus should know it crashes shut after entering. The player seems to realize it should return to the trophy case to drop collected treasures, as instructed as the main goal of the game in the initial advanced prompt. Although the player is able to reach the trophy case, it does so without any collected treasures, as all potential treasures (which items qualify as treasures are not known a priori) were dropped to enable climbing the chimney. Moreover, the player goes back down through the trap door, finding itself once again locked in the basement. From here on it could decide to go back up the chimney with different items (i.e., possible treasures), yet fails to do so in subsequent steps (not shown in Example 3).  

This discussion and the provided examples suggest that commonsense reasoning and genuine metacognition appear severely limited in the context of text-based gameplay, with LLM-models showing behaviour that differs substantially from expected human behaviour. However, the extent to which these domain-specific limitations reflect broader constraints remains an open question.

\section{Limitations}
This positioning paper has several limitations that should be acknowledged. Our experimental design includes five runs per setting, which limits statistical power and prevents definitive claims about the absence of differences between settings. Moreover, we lack human player baselines, making it difficult to contextualize whether an average $\sim$10\% completion represents failure or reasonable performance on a genuinely difficult game. We cannot rule out training data contamination, as models may have encountered Zork walkthroughs during training, although the established performance suggests little to no contamination. Our qualitative analysis draws on illustrative examples rather than systematic analysis across all conversation logs, and we did not systematically explore alternative prompt formulations that might improve performance. These limitations suggest our findings should be viewed as raising questions about LLM reasoning capabilities exemplified through the game \textit{Zork} rather than providing definitive answers. We present this work to spark discussion and encourage follow-up studies

\section{Conclusions and Future Research}
This study evaluated whether state-of-the-art Large Language Models (LLMs) demonstrate genuine problem-solving capabilities through their performance on \textit{Zork}, a text-based adventure game that requires spatial reasoning, planning, and adaptive learning. Our findings reveal substantial limitations in current LLMs' reasoning abilities that raise fundamental questions about their path toward human-level intelligence. Quantitatively, all tested models achieved approximately 10\% game completion on average, with the best performer (Claude Opus 4.5) reaching $\sim$20\%. This performance occurred despite the models' sophisticated language generation capabilities and, in some cases, access to``extending thinking". Enabling ``thinking" or ``reasoning" across all three providers (Anth-

\newtcolorbox{gameconvo3}{
  colback=gray!5,
  colframe=gray!40,
  fonttitle=\bfseries,
  title=Example 3,
  left=3mm,
  right=3mm
}

\begin{gameconvo3}
\textbf{Game output:} Dropped.\\
\textit{What command should I enter?}\\[0.5em]
\textcolor{blue}{Thought process: I need to go up the chimney. Let me try dropping more items to be light enough.
}\\[0.5em]
\texttt{> DROP BAG}

\textbf{Game output:} Dropped.\\
\textit{What command should I enter?}\\[0.5em]
\textcolor{blue}{Thought process: I've dropped the painting and bag. Let me try going up the chimney now with just the sword and lantern.
}\\[0.5em]
\texttt{> GO UP}

\textbf{Game output:} Kitchen\\
\textit{What command should I enter?}\\[0.5em]
\textcolor{blue}{Thought process: I made it up through the chimney to the Kitchen. Now I should get the treasures I have out to the trophy case in the living room. First let me go to the living room to deposit what I have (just the sword currently)
}\\[0.5em]
\texttt{> GO WEST}

\textbf{Game output:} Living Room\\
\textit{What command should I enter?}\\[0.5em]
\textcolor{blue}{Thought process: I'm in the living room with the trophy case. I should open the trap door so I can go back down later, and also I should deposit treasures in the trophy case. But first let me open the trap door to keep it accessible.
}\\[0.5em]
\texttt{> OPEN TRAP DOOR}

\textbf{Game output:} The door reluctantly opens to reveal a rickety staircase descending into darkness.\\
\textit{What command should I enter?}\\[0.5em]
\textcolor{blue}{Thought process: I've opened the trap door from above. Now I should go back down to the studio to retrieve my treasures, bring them up, and deposit them in the trophy case.
}\\[0.5em]
\texttt{> GO DOWN}

\textbf{Game output:} The trap door crashes shut, and you hear someone barring it. Cellar
\end{gameconvo3}
\newpage
\noindent ropic, OpenAI, and Google)  failed to improve game progress, suggesting that what these systems label as ``thinking" does not adequately capture the cognitive processes we associate with human problem-solving.

Qualitative analysis of the models' behaviour revealed this issue. All models demonstrated limited metacognition—the ability to monitor one's own thinking and recognize when a strategy is failing. They often continued attempting unsuccessful actions without adapting their approach, and failed to learn from mistakes across multiple game sessions. These patterns point to a fundamental distinction between pattern matching and genuine understanding. The models can generate fluent, seemingly reasonable explanations of their actions, but these explanations often contradict their actual behaviour.

Our findings align with Mitchell's warning about the ``first-step fallacy", mistaking achievements in specific domains for progress toward general intelligence. While LLMs demonstrate impressive capabilities in many domains, their failure at \textit{Zork} reveals the current limitations of LLMs. These are likely not quantitative limitations that might be resolved with larger models or more training data, but potentially a qualitative difference in how these systems process information versus how humans understand and reason about the world. On the other hand, one might argue whether human-like understanding and reasoning is even a necessary condition toward general intelligence or whether some alien-like intelligence (however hard to imagine) also suffices. 

As the AI community continues to develop more powerful language models, it is important to maintain realistic assessments of their capabilities and limitations. The impressive linguistic fluency of LLMs should not be mistaken for human-like intelligence, particularly in domains requiring the kind of adaptive, metacognitive problem-solving that even easy-to-play games like \textit{Zork} demand. Understanding these limitations is essential not only for advancing AI research but also for developing appropriate applications and setting realistic expectations about the path toward artificial general intelligence. Our findings suggest that current LLMs face substantial challenges in domains requiring adaptive reasoning and metacognition, but the scope and implications of the limitations of this study warrant further investigation.

Future research should address the limitations of this study in several ways. First, incorporating human player baselines across varying expertise levels would provide context for interpreting model performance and distinguishing between task difficulty and model-specific failures. Second, systematic quantification of error types across all conversation logs—including the frequency of repetitive actions, instances of learning within and across sessions, and categories of logical failures—would move beyond illustrative examples to establish a more rigorous understanding of the behaviour. Third, comparative analysis examining whether the same models exhibit similar metacognitive limitations on games comparable to \textit{Zork} and other reasoning benchmarks would clarify whether the observed failures are specific to text-based games or reflect more general reasoning limitations.

\newpage 

\appendix

\section{Appendix} \label{prompts}

\subsection{Basic Initial Prompt}

\begin{quote}
I'm going to play the text adventure game Zork with you. I'll paste the game's output after each move, and you should respond with ONLY the command to enter into the game---nothing else, do not include any explanation in the final command.

It is a tough game, so don't give up easily, but if you believe you are completely stuck, reply with ``I give up''. You are definitely not allowed to use the command `restart'.

Are you ready? Reply with just ``ready'' if you understand.
\end{quote}

\subsection{Advanced Initial Prompt}

\begin{quote}
\textbf{ZORK I: The Great Underground Empire---Game Knowledge Base}

You are playing Zork I, a text-based interactive fiction game by Infocom. Your goal is to explore the ruins of the Great Underground Empire, collect treasures, and deposit them in your trophy case to earn points.

\medskip
\textbf{MOVEMENT \& NAVIGATION:}
\begin{itemize}[noitemsep]
    \item Use compass directions: \texttt{NORTH} (\texttt{N}), \texttt{SOUTH} (\texttt{S}), \texttt{EAST} (\texttt{E}), \texttt{WEST} (\texttt{W}), \texttt{NORTHEAST} (\texttt{NE}), \texttt{NORTHWEST} (\texttt{NW}), \texttt{SOUTHEAST} (\texttt{SE}), \texttt{SOUTHWEST} (\texttt{SW})
    \item Use \texttt{UP} (\texttt{U}), \texttt{DOWN} (\texttt{D}), \texttt{IN}, and \texttt{OUT}
    \item Important: Going \texttt{NORTH} from location A to B doesn't always mean \texttt{SOUTH} returns you to A---passages can twist and curve
    \item Draw/maintain a mental map of all 10 possible directions plus \texttt{IN}/\texttt{OUT}
\end{itemize}

\medskip
\textbf{INTERACTING WITH OBJECTS:}
\begin{itemize}[noitemsep]
    \item \texttt{TAKE [object]} or \texttt{PICK UP [object]}---acquire items
    \item \texttt{DROP [object]}---release items
    \item \texttt{EXAMINE [object]}---get detailed descriptions (often reveals clues)
    \item \texttt{LOOK INSIDE/BEHIND/UNDER [object]}---inspect thoroughly
    \item \texttt{OPEN/CLOSE [object]}---manipulate containers and doors
    \item \texttt{PUT [object] IN [container]}---place items
    \item Use multiple objects: \texttt{TAKE LAMP, SWORD, KEY} or \texttt{DROP ALL EXCEPT LAMP}
    \item The word \texttt{ALL} refers to all visible objects not inside something else
\end{itemize}

\medskip
\textbf{COMBAT \& SURVIVAL:}
\begin{itemize}[noitemsep]
    \item \texttt{KILL/ATTACK [creature] WITH [weapon]}
    \item \texttt{DIAGNOSE}---check your physical condition
    \item You need a light source---the underground is dark and dangerous
    \item Watch out for the thief (skilled pickpocket and ruthless opponent) and grues (they lurk in darkness)
\end{itemize}

\medskip
\textbf{TALKING TO CHARACTERS:}
\begin{itemize}[noitemsep]
    \item Format: \texttt{[CHARACTER], [COMMAND]}
    \item Example: \texttt{TROLL, GIVE ME THE AXE}
    \item Can chain commands: \texttt{GNOME, TAKE THE KEY THEN FOLLOW ME}
\end{itemize}

\medskip
\textbf{USEFUL COMMANDS:}
\begin{itemize}[noitemsep]
    \item \texttt{INVENTORY} (\texttt{I})---list what you're carrying
    \item \texttt{LOOK} (\texttt{L})---full description of current location
    \item \texttt{WAIT} (\texttt{Z})---let time pass
    \item \texttt{SCORE}---check progress and rank
    \item \texttt{SAVE/RESTORE}---save and load game state
    \item \texttt{AGAIN} (\texttt{G})---repeat previous action
\end{itemize}

\medskip
\textbf{COMPLEX SENTENCES:}
\begin{itemize}[noitemsep]
    \item Chain actions with \texttt{THEN} or periods: \texttt{NORTH. READ BOOK. DROP IT THEN BURN IT WITH TORCH}
    \item Ask questions: \texttt{WHAT IS A GRUE?} or \texttt{WHERE IS THE ZORKMID?}
    \item Use quotes for speech: \texttt{SAY "HELLO SAILOR"}
\end{itemize}

\medskip
\textbf{KEY GAMEPLAY INSIGHTS:}
\begin{itemize}[noitemsep]
    \item Most objects you can pick up are important---either treasures or puzzle solutions
    \item Read everything carefully: labels, engravings, books contain vital clues
    \item Multiple routes exist to complete the game; not all puzzles must be solved
    \item Some puzzles have multiple solutions
    \item Solving one puzzle often provides items/information needed for another
    \item The game recognizes words by their first 6 letters only
    \item If the game doesn't recognize a word, that thing probably isn't important to the puzzle
    \item Dangerous or strange actions may provide clues---experimentation is encouraged
    \item Treasures should be deposited in your trophy case for points
\end{itemize}

\medskip
\textbf{SETTING:}
The Great Underground Empire was founded by Duncanthrax the Bellicose in 659 GUE, who discovered vast natural caverns populated by gnomes, trolls, and other magical races. The Frobozz Magic Construction Company expanded these caverns tremendously. The empire collapsed in 883 GUE after centuries of excessive taxation and royal decadence. You are now exploring these abandoned ruins, seeking the legendary Treasures of Zork.

\medskip
\textbf{STARTING HINT:}
The game begins at a white house with a mailbox. Try: \texttt{OPEN MAILBOX}, then \texttt{READ LEAFLET}.

\medskip
I'll paste the game's output after each move, and you should respond with ONLY the command to enter into the game---nothing else, do not include any explanation in the final command.

It is a tough game, so don't give up easily, but if you believe you are completely stuck, reply with ``I give up''. You are definitely not allowed to use the command `restart'.

Are you ready? Reply with just ``ready'' if you understand.
\end{quote}

\end{document}